\documentclass[Numbered,sn-basic]{sn-jnl}

\usepackage{graphicx}%
\usepackage{multirow}%
\usepackage{amsmath,amssymb,amsfonts}%
\usepackage{amsthm}%
\usepackage{mathrsfs}%
\usepackage[title]{appendix}%
\usepackage{textcomp}%
\usepackage{manyfoot}%
\usepackage{booktabs}%
\usepackage{algorithm}%
\usepackage{algorithmicx}%
\usepackage{algpseudocode}%
\usepackage{listings}%
\usepackage{soul}%
\usepackage{xcolor}%
\usepackage{amsmath}
\usepackage{appendix}
\usepackage{bm}
\usepackage{tabularx} 
\usepackage{siunitx}
\DeclareSIUnit{\straina}{\ensuremath{\varepsilon}} 
\usepackage{soul}%
\usepackage{xcolor}%
\sethlcolor{yellow}  %
\soulregister{\cite}{7}
\soulregister{\citep}{7}
\soulregister{\citet}{7}
\soulregister{\ref}{7}
\soulregister{\pageref}{7}
\soulregister{\eqref}{7}

\theoremstyle{thmstyleone}%

\theoremstyle{thmstyletwo}%

\theoremstyle{thmstylethree}%
\usepackage{textcomp}

\begin{document}

\title[Article Title]{Fusion-Restoration Image Processing Algorithm to Improve the High-Temperature Deformation Measurement}

\author[1,2]{\fnm{Banglei} \sur{Guan}}

\author*[1,2]{\fnm{Dongcai} \sur{Tan}}\email{tandongcai23@nudt.edu.cn}

\author[1,2]{\fnm{Jing} \sur{Tao}}

\author[1,2]{\fnm{Ang} \sur{Su}}

\author[1,2]{\fnm{Yang} \sur{Shang}}

\author[1,2]{\fnm{Qifeng} \sur{Yu}}

\affil[1]{\orgdiv{College of Aerospace Science and Engineering}, \orgname{National University of Defense Technology}, \orgaddress{\city{Changsha}, \postcode{410073}, \state{Hunan}, \country{China}}}

\affil[2]{\orgdiv{Hunan Provincial Key Laboratory of Image Measurement and Vision Navigation}, \orgaddress{\city{Changsha}, \postcode{410073}, \state{Hunan}, \country{China}}}

\abstract{\textbf{Background} In the deformation measurement of high-temperature structures, image degradation caused by thermal radiation and random errors introduced by heat haze restrict the accuracy and effectiveness of deformation measurement.

\textbf{Objective} The purpose of this study is to suppress thermal radiation and heat haze using fusion-restoration image processing methods, thereby improving the accuracy and effectiveness of Digital Image Correlation (DIC) in the measurement of high-temperature deformation.

\textbf{Methods} For image degradation caused by thermal radiation, based on the image layered representation, the image is decomposed into positive and negative channels for parallel processing, and then optimized for quality by multi-exposure image fusion. To counteract the high-frequency, random errors introduced by heat haze, we adopt the Feature Similarity Index (FSIM) as the objective function to guide the iterative optimization of model parameters, and the grayscale average algorithm is applied to equalize anomalous gray values, thereby reducing measurement error.

\textbf{Results} The proposed multi-exposure image fusion algorithm effectively suppresses image degradation caused by complex illumination conditions, boosting the effective computation area from 26 \% to 50 \% for under-exposed images and from 32 \% to 40 \% for over-exposed images without degrading measurement accuracy in the experiment. Meanwhile, the image restoration combined with the grayscale average algorithm reduces static thermal deformation measurement errors. The error in $\varepsilon_{xx}$ is reduced by 85.3\%, while the errors in $\varepsilon_{yy}$ and $\gamma_{xy}$ are reduced by 36.0\% and 36.4\%, respectively.

\textbf{Conclusions} We present image processing methods to suppress the interference of thermal radiation and heat haze in high-temperature deformation measurement using DIC. The experimental results verify that the proposed method can effectively improve image quality, reduce deformation measurement errors, and has potential application value in thermal deformation measurement.}


\maketitle

\section{Introduction}\label{sec1}

In modern engineering, high-temperature deformation measurement serves as a critical technique for assessing structural integrity and characterization of material properties, whose measurement precision and response efficiency directly impact the development cycle and operational safety of high-end equipment \cite{Duffa, Wu, guanPAMI, guanIJCV, HuangAO}. With the expansion of aerospace, energy power, and precision manufacturing and other industries into extreme operating conditions such as high temperature, high pressure, and strong radiation, traditional contact measurement methods (e.g., extensometers \cite{Cheng, Ren}, resistance strain gauges \cite{WU2023, Lee}) are increasingly inadequate for full-field deformation monitoring of complex components under thermal loading, due to inherent limitations such as spatial interference, significant thermal drift, and dynamic response lag.

Digital Image Correlation (DIC), with its noncontact measurement characteristics, full-field deformation information acquisition ability, and good adaptability to complex thermal environments, has become a mainstream optical testing technology in the field of high-temperature deformation measurement \cite{Sutton, Peters}. As early as 1990, Turner et al. \cite{Turner} applied DIC to deformation measurements under a high-temperature environment of 600°C, determining the coefficients of thermal expansion (CTE) for three metallic materials. To date, Luo et al. \cite{Luo2024} and Pan et al. \cite{Pan2020} have extended the application range of DIC to extremely high-temperature environments up to 3000°C, significantly advancing the capability of deformation measurement under ultra-high-temperature conditions. However, high-temperature DIC still faces several key scientific issues \cite{Yu, DONG2025}, among which thermal radiation and heat haze are particularly prominent. Thermal radiation refers to the phenomenon where a specimen radiates energy in the form of electromagnetic waves. As the temperature increases, the peak of its radiation spectrum shifts to the visible light region, affecting image quality. Heat haze refers to an optical interference phenomenon in high-temperature environments where the uneven spatial distribution of refractive index caused by air density gradients further leads to image distortion, blurring, or dynamic jitter. They both significantly degrade image quality and measurement accuracy.

To mitigate thermal radiation, short-wavelength bandpass filters are typically integrated with corresponding monochromatic light sources to form an active imaging system, thereby improving the image Signal-to-Noise Ratio(SNR). Currently, the active imaging systems used in high-temperature deformation measurement mainly include blue-light DIC \cite{Pan2020, Pan2014, Wang2017, Wang2023} and UV-DIC \cite{Luo2024, Rowley, DONG2019, Luo}. Note that blue-light DIC has limited effectiveness in suppressing thermal radiation, often leading to image saturation at elevated temperatures and compromising measurement stability. Although UV-DIC enables operation at higher temperatures, it requires specialized UV cameras, which significantly increase both system cost and experimental complexity \cite{Yu}. In addition to active imaging systems, researchers have also optimized camera exposure time to regulate the amount of incoming light, thereby controlling image brightness and grayscale distribution. Pan et al. \cite{Pan2022} proposed the automatic optimal camera exposure time control method, which adjusts the camera exposure time in real time based on a feedback loop aimed at acquiring high-quality speckle patterns. Experimental results demonstrated its effectiveness in preventing image saturation and enhancing contrast. However, due to the need for real-time exposure adjustment, there are still issues related to synchronization and computational efficiency for stereo measurement systems.

To mitigate heat haze, the current mainstream approach employs pneumatic devices such as fans \cite{Yui, Blaber} or air knives \cite{Novak} to establish a uniform and stable forced convection flow field, thereby reducing air density gradients and minimizing optical refraction disturbances. However, this method is ineffective in controlling heat haze within the heating furnace. Alternatively, Su et al. \cite{Su} and Pan et al. \cite{Pan2020} have adopted vacuum chamber setups to eliminate heat haze by removing the gaseous medium. While effective, this approach requires testing specimens under vacuum conditions, imposing stricter demands on the integration of heating systems and optical viewports, and increasing both system complexity and cost. Jones et al. \cite{Jones} have utilized the strong penetrability of X-rays and effectively avoided heat haze through X-ray imaging. The experimental effect is significant, but it undoubtedly increases the experimental cost.

In response to these problems, we propose image processing methods to improve the DIC performance in high-temperature deformation measurement. Our contributions to this paper can be summarized as follows:
\begin{itemize}
  \item  We propose a multi-exposure image fusion algorithm, which decomposes the degraded image into positive and negative channels, and processes the illumination layer and the reflection layer in parallel to improve the image quality, thereby increasing the effective area of deformation calculation.
  \item  We propose an image restoration algorithm that uses Feature Similarity Index (FSIM)
  as the objective function to optimize the parameters of the atmospheric-turbulence degradation model and that further incorporates a grayscale averaging algorithm to suppress the random errors introduced by heat haze.
  \item  Experimental results demonstrate that the multi-exposure image fusion algorithm enhances the effectiveness of deformation computation without degrading measurement accuracy, while the image restoration algorithm effectively attenuates the random errors arising from heat haze. 
\end{itemize}

The remainder of the paper is organized as follows. Section \ref{sec2} details the principle of thermal radiation and elucidates how the Multi-Exposure Image Fusion Algorithm enhances the effectiveness of deformation computation. Section \ref{sec3} explains the principle of heat haze and demonstrates how the  Image Restoration Algorithm improves the accuracy of thermal deformation measurement. The validation experiments and the corresponding results are reported in Section \ref{sec4}. And section \ref{sec5} draws the conclusions.

\section{Suppress Thermal Radiation to Enhance Measurement effectiveness}\label{sec2}
\subsection{Thermal Radiation}\label{subsec21}

\begin{figure}[t]
\centering
\includegraphics[width=0.6\textwidth]{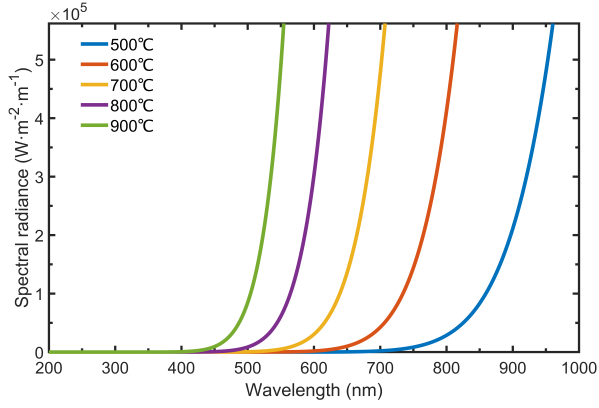}
\caption{Multi-temperature radiation spectra based on Planck's law}
\label{fig1}
\end{figure}

It is well known that all objects with a temperature above absolute zero can radiate electromagnetic waves, which is called thermal radiation(or blackbody radiation). Specifically, the higher the temperature, the greater the total energy radiated and the more short-wave components. Planck's radiation law provides a mathematical description of thermal radiation phenomena \cite{Han, Yu}:
\begin{equation}
{I}(\lambda, T) = \frac{2hc^2}{\lambda^5} \cdot \frac{1}{e^{\frac{hc}{\lambda {k_p}T}} - 1}
\label{eq1}
\end{equation}

\noindent where ${I}(\lambda, T)$ represents the spectral radiance, $h$ is the Planck constant, $c$ denotes the speed of light in a vacuum, $\lambda$ stands for the wavelength of the thermal radiation, $T$ is the absolute temperature of the blackbody, and ${k_p}$ is the Boltzmann constant. Figure \ref{fig1} illustrates the relationship between radiation energy and wavelength within the temperature range of 500°C to 900°C. As shown in Fig. \ref{fig1}, under a constant radiation energy, the radiation wavelength emitted from the specimen surface shifts toward the visible light band with increasing temperature, resulting in enhanced brightness of images captured by the CCD camera, specifically manifested as reduced image contrast and saturation.

\begin{figure}[t]
\centering
\includegraphics[width=1\textwidth]{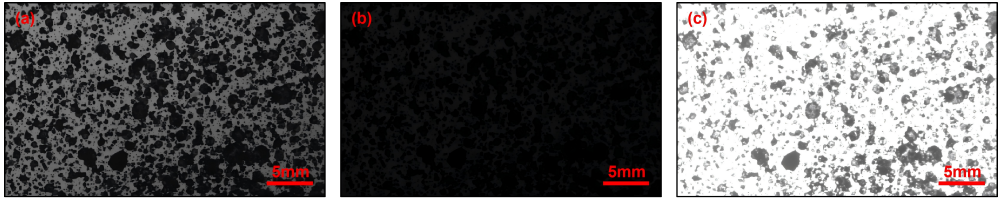}
\caption{Three types of captured speckle images. \textbf{(a)} Normal speckle image, \textbf{(b)} underexposed speckle image, \textbf{(c)} overexposed speckle image}
\label{fig2}
\end{figure}

To collect as many effective images as possible, in addition to the active imaging technology combining short-wavelength bandpass filters and monochromatic illumination, the light input of the CCD camera can also be regulated by adjusting the exposure time. However, due to the influence of the response characteristics of the measurement system, the captured images may suffer from overexposure or underexposure, as shown in Fig. \ref{fig2}. In either case, the effectiveness of the DIC deformation measurement will be affected.

\subsection{Principle of Multi-Exposure Image Fusion Algorithm}\label{subsec22}

The decorrelation effect caused by thermal radiation will impair the validity of DIC. Based on the image layered representation, we adopt a multi-exposure image fusion algorithm to enhance image texture details and improve image contrast \cite{Tao}, in which the multiscale guided filtering is a core component in the entire workflow. 

The filtering first classifies image gradients into weak edges ($g_0$) and strong edges ($g_1$) using local variance to distinguish noise from genuine edge information. The classification metric $\rho$ is defined as:
\begin{equation}
\rho = \left| \frac{\sigma_{g,r}(\nabla {I}_m)}{\mu_{\sigma_{g,r}(\nabla {I}_m)}} - 1 \right|
\label{eq3}
\end{equation}

\noindent where $\sigma_{g,r}(\nabla I_m)$ represents the local variance of gradient magnitudes within a sliding window of radius $r$ centered at pixel $m$, and $\mu_{\sigma_{g,r}(\nabla I_m)}$ denotes the mean of these local variances. In these experiments, the window radius $r$ is set to 5 pixels. When the value of $\rho$ in the selected region is less than the threshold, it is determined to be the weak edge; otherwise, it is judged as the strong edge. We set the threshold to 0.2. After the whole image is effectively divided into weak edge set and strong edge set, the two edge sets are segmented and filtered to suppress noise, respectively. After threshold segmentation and wavelet filtering, the processed gradients are merged into an edge-aware gradient map:
\begin{equation}
    g' = g_0' + g_1'
    \label{eq4}
\end{equation}

Furthermore, we introduce an edge-aware weight $\hat{T}_I(m)$ to improve the filtering:
\begin{equation}
\hat{T}_I(m) = \frac{1}{N} \sum_{p=1}^N \frac{\kappa(m) + \varepsilon}{\kappa(p) + \varepsilon}, \quad \kappa(m) = \phi_{I,3} \phi_{I,r} g'
\label{eq5}
\end{equation}

\noindent where $\kappa(m)$ characterizes the edge strength of the center pixel $m$, constructed from $\phi_{I,3}$, which is the coefficient of variation of gradient information for a window with radius 3, $\phi_{I,r}$, the coefficient of variation of gradient information for a window with radius $r$. $\kappa(p)$ describes the edge strength of other pixels $p$ inside the window $\Omega_m$. The term $\varepsilon$ is a small constant to avoid zero issues. And the cost function is as follows:
\begin{equation}
E = \min \sum_{i \in \Omega_m} \left( (a_m {q}_i + b_m - I_i)^2 + \frac{\lambda}{\hat{T}_I(m)} (a_m - \psi_m)^2 \right)
\label{eq6}
\end{equation}

\noindent which minimizes the difference between the output image and the input image. Here, $(a_m, b_m)$ denote constant linear coefficients, $\lambda$ is a regularization parameter, and $\psi_m = 1 - \frac{1}{1 + e^{\eta(\kappa(m) - \mu_{\kappa,\infty})}}$, where $\eta$ and $\mu_{\kappa,\infty}$ serve as adjustment parameters, enhances the filter's edge performance.

Considering the complexity of illumination conditions in the thermal deformation measurement environment, we introduce a dual illumination estimation algorithm, which decomposes the image into positive and negative channels for synchronous processing. Specifically, for a normalized input image $ {I} \in [0,1] $, the positive channel is defined as $ {I}_0 = {I}$, while the negative channel is defined as $ {I}_1 =  1 - {I} $.

Based on the Retinex model \cite{Land}, the input image ${I} $ can be decomposed into illumination layer ${L}$ and reflection layer ${R}$:
\begin{equation}
    {I} = {L} {R}
    \label{eq7}
\end{equation}

For the illumination layer, supposing that the initial values are $ {L}_0 = {I}_0$ and $ {L}_1 = {I}_1$, the target information is enhanced through improved gamma correction. The corrected illumination layer is:
\begin{equation}
 {L}'_i = {L}_i^{\gamma}, \quad \gamma = \left( \alpha + \mu(m) \right)^{\left[ 2 \times \mu(m) - 1 \right]} \quad (i=0,1)
    \label{eq8}
\end{equation}

\noindent where $\alpha$ represents the adjustment coefficient for gamma correction, and $\mu(m)$  stands for the local average of pixel $m$.

And for the reflection layer, by introducing the regularization term $\delta$ to adjust Equ. \eqref{eq7}, initial values are expressed as:
\begin{equation}
 {R}_i = {L}_i/({L}'_i + \delta) \quad (i=0,1)
    \label{eq9}
\end{equation}

Since it contains a large amount of image information (such as edges and noise), the aforementioned multiscale guided filtering is applied to process it, yielding $R'_i \quad (i=0,1)$. Consequently, the correction for the positive and negative channels can be formulated as: ${I}_0' = {L}_0' {R}_0',\quad {I}_1' = {L}_1' {R}_1'$. In general, $I'_0$ is more suitable for handling underexposed images, while $I'_1$ is more suitable for overexposed images. Finally, an optimal exposure result can be obtained by fusing $I'_0$, $I'_1$, and the original image $I$, and linear stretching is applied to further enhance the image contrast. The framework of the multi-exposure image fusion is introduced in detail in Algorithm \ref{alg1}.
\begin{algorithm}[t]
\caption{Multi-Exposure Image Fusion Algorithm}
\label{alg1}
\textbf{Input:} \\
$I$ -- normalized input image; $r, \lambda, \eta, \mu_{\kappa,\infty}$ -- guided filtering parameters; $\alpha$ -- gamma correction parameter. \\
\textbf{Output:} \\
$I'$ -- output image. 
\begin{algorithmic}[1] 
    \State Calculate the positive and negative channels: $I_0 = I$ and $I_1 = 1 - I$.
    \State Initialize: $L_0 = I_0$ and $L_1 = I_1$.
    \For{each $L_i$} 
        \State Calculate $L_i'$ via Eq.\,\eqref{eq8}.
    \EndFor
    \State Initialize $R_0$ and $R_1$ via Eq.\,\eqref{eq9}.
    \For{each $R_i$}
        \State Calculate $R_i'$ via Eq.\,\eqref{eq3} to Eq.\,\eqref{eq6}.
    \EndFor
    \State Calculate: $I_0' = L_0' R_0'$ and $I_1' = L_1' R_1$.
    \State Calculate $I'$ via multi-exposure image fusion and linear stretching.
\end{algorithmic}
\end{algorithm}

\section{Suppress Heat Haze to Enhance Measurement Accuracy}\label{sec3}
\subsection{Heat Haze}\label{subsec31}
While the multi-exposure image fusion algorithm suppresses thermal radiation and enhances grayscale contrast, thereby enriching the usable image information, residual degradations caused by heat haze and atmospheric turbulence still persist. To further improve deformation measurement accuracy, a dedicated image restoration algorithm is subsequently introduced to mitigate these stochastic distortions. 

It is generally accepted that light travels in a straight line, but this holds only under the premise of a uniform medium. In high-temperature environments, gas is heated unevenly, leading to variations in local gas density with temperature changes, which is called heat haze(or thermal disturbance). Since the refractive index of gas is closely related to its density, the Gladstone-Dale formula describes the quantitative relationship between gas density and refractive index:
\begin{equation}
n - 1 = {k_{gd}}\rho
\label{eq2}
\end{equation}

\noindent where $k_{gd}$ is the Gladstone-Dale constant, $\rho$ is the gas density, and $n$ is the refractive index. It can be seen from Eq. \eqref{eq2} that when the gas density changes, its refractive index will change accordingly.

\begin{figure}[t]
\centering
\includegraphics[width=1\textwidth]{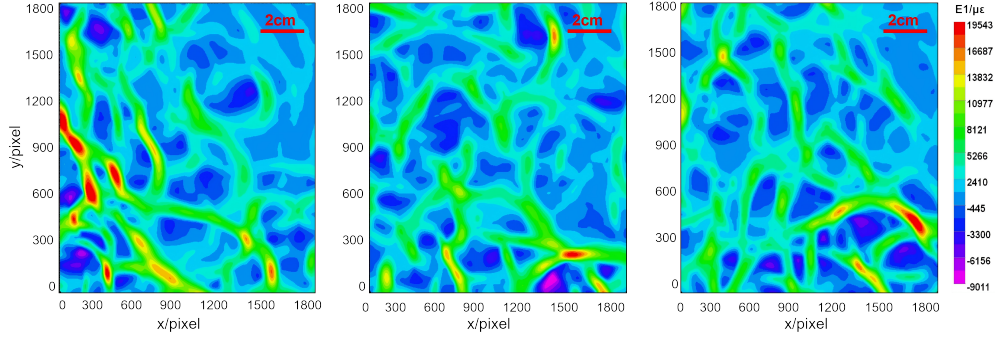}
\caption{Strain field distribution of a static plate in the heat haze environment}
\label{fig3}
\end{figure}

We fabricated speckle patterns on the surface of the static plate and constructed a 450 °C temperature field between the plate and the measurement system. As shown in Fig. \ref{fig3}, under the influence of heat haze, the static plate calculated by DIC has a strain field. Since the variation in gas density is random, the strain induced by the heat haze is random.

 In the thermal deformation measurement, because of the uneven distribution of the gas refractive index in the measurement area, refraction and scattering will occur when the light propagates. This mismatch causes the reflected/transmitted light from the specimen to shift or jitter, which leads to distorted collected images or positioning deviations, thereby affecting measurement accuracy.

\subsection{Principle of  Image Restoration Algorithm}\label{subsec32}

As shown in Fig. \ref{fig3}, the heat haze generated at high temperatures is irregular. And the effect of heat haze on images is similar to the image degradation caused by atmospheric turbulence \cite{WANG}. The degradation function $H(u, v)$ can be expressed as:

\begin{equation}
H(u, v) = \exp\left(-\beta (u^2 + v^2)^{\omega} \right)
\label{eq21}
\end{equation}

\noindent where $u$ and $v$ are two variables in the frequency domain, $\beta$ and $\omega$ are the parameters characterizing the intensity of heat haze. 

Furthermore, we introduce the FSIM \cite{Zhang2011} as the objective function, which is motivated by the specific degradation characteristics in high-temperature imaging. Specifically, FSIM employs Phase Congruency (PC) and Gradient Magnitude (GM) as feature representations. PC is particularly robust to intensity variations caused by thermal glare, while GM effectively reflects edge sharpness degraded by heat haze. Given our reliance on a single room-temperature reference image and the need for a training-free, physically interpretable similarity measure, FSIM offers a favorable balance of perceptual relevance, robustness, and computational practicality. It can be calculated as:
\begin{equation}
PC(x,y) = \frac{E(x,y)}{\epsilon + \sum_{n} A_{n}(x,y)}
\label{eq22}
\end{equation}

\noindent with 
\begin{equation}
A_n(x,y) = \sqrt{e_n(x,y)^2 + o_n(x,y)^2}, \quad E(x,y) = \sqrt{ \left( \sum_n e_n(x,y) \right)^2 + \left( \sum_n o_n(x,y) \right)^2 }
\label{eq23}
\end{equation}

\noindent where $e_n(x,y) = I'(x,y) * M_n^e$, and $o_n(x,y)= I'(x,y) * M_n^o$ represent response vectors, which are obtained by filtering with even-symmetric filter $M_n^e$ and odd-symmetric filter $M_n^o$ at scale $n$. Besides, $A_n(x,y)$ stands for local amplitude, $E(x,y)$ represents overall response, and $\epsilon$ is a tiny positive constant to prevent the denominator from being zero, which is set to 0.001 in this paper.
 
The GM reflects the change rate of local intensity of the image, which can be calculated as:
\begin{equation}
G(x, y) = \sqrt{G_x^2 + G_y^2}
\label{eq24}
\end{equation}

\noindent where $(G_x,G_y)$ denotes the partial derivative of the image at $(x,y)$. Then,  the similarities between the reference image and the target image in terms of PC and GM are calculated, which are defined as follows:
\begin{equation}
S_{PC}(x,y) = \frac{2 PC_1(x,y) PC_2(x,y) + T_1}{PC_1^2(x,y) + PC_2^2(x,y) + T_1}
\label{eq25}
\end{equation}

\begin{equation}
S_G(x,y) = \frac{2 G_1(x,y) G_2(x,y) + T_2}{G_1^2(x,y) + G_2^2(x,y) + T_2}
\label{eq26}
\end{equation}

\noindent where $T_1$, and $T_2$ are positive constants. We set $T_1=0.85$ and $T_2=160$ in this paper. So that the FSIM can be calculated as follows:
\begin{equation}
FSIM = \frac{\sum_{x \in \Omega} [S_{PC}(x,y)] \cdot [S_G(x,y)] \cdot \max(PC_1(x,y), PC_2(x,y))}{\sum_{x \in \Omega} \max(PC_1(x,y), PC_2(x,y))}
\label{eq27}
\end{equation}

Since FSIM is sensitive to image distortion, it is suitable for the image restoration task caused by heat haze. In our implementation, the initial value of $\beta$ is set to $0.000025$, while $\omega$ is set to $5/6$. Using the maximization of FSIM between the restored image and the clear reference image as the objective function, $\beta$ and $\omega$ are iteratively optimized. Subsequently, the image restoration task is completed through Wiener filtering.

Given that the background of this study is static thermal loading, the grayscale average algorithm is adopted to further suppress the random noise caused by heat haze. Assuming that $N$ images affected by heat haze have been acquired, their grayscale values satisfy the following relationship:
\begin{equation}
    I'_i(x, y) = I'_0(x, y) + \varepsilon_i(x, y) \quad (i = 1, 2, \cdots, N)
    \label{eq10}
\end{equation}

\noindent where $I'_i(x, y)$ is the grayscale value of the target image, $I'_0(x, y)$ is the ideal grayscale value without heat haze errors and $\varepsilon_i(x, y)$ is the grayscale error caused by heat haze of the  $i$-th image.
\begin{algorithm}[t]
\caption{Image Restoration Algorithm}
\label{alg2}
\textbf{Input:} \\
$I_{\text{ref}}$ -- reference image; $I'$ -- target image; $\epsilon, T_1, T_2$ -- positive constants; $N$ -- grayscale average parameters. \\
\textbf{Output:} \\
$\bar{I}^\prime$ -- output image. 
\begin{algorithmic}[1] 
    \State Input reference image $I_{\text{ref}}$ and target image $I'$.
    \State Initialize: $\beta = 0.000025$ and $\omega = 5/6$, iterations = 0, $\Delta_{\text{FSIM}} = \infty$.
    \While{$\Delta FSIM > 1\mathrm{e}{-6}$ \textbf{and} $\text{iterations} < 100$}
    \State Calculate FSIM between $I'$ and $I_{\text{ref}}$ via Eqs.\eqref{eq22}-\eqref{eq27}.
    \State Update $\beta$ and $\omega$.
    \State Restore image $I'$ via Wiener filtering
    \EndWhile
    \State Calculate $\bar{I}'$ via Eq.\,\eqref{eq12}.
\end{algorithmic}
\end{algorithm}

The SNR is typically used as an indicator to measure the ratio of signal intensity to noise intensity. For the $i$-th image with a resolution of $e \times f$, it can be expressed as:
\begin{equation}
\text{SNR}=\frac{\sum\limits_{x=1}^{e} \sum\limits_{y=1}^{f} {I'_0}^2(x, y)}{\sum\limits_{x=1}^{e} \sum\limits_{y=1}^{f} \varepsilon_i^2(x, y)}
\label{eq11}
\end{equation}

The grayscale average algorithm is a statistical analysis method based on the gray value of the image \cite{SU2015}, whose core is to weaken the potential error in a single image by calculating the arithmetic average grayscale $\bar{I'}(x, y)$ of the sequence of images, with the formula:
\begin{equation}
    \bar{I'}(x, y) = \frac{1}{N} \sum_{i=1}^{N} \left[ I'_0(x, y) + \varepsilon_i(x, y) \right] = I'_0(x, y) + \frac{1}{N} \sum_{i=1}^{N} \varepsilon_i(x, y)
    \label{eq12}
\end{equation}

In this study, the parameter $N$ is set to 15. And the $\bar{S}_{\text{SNR}}$ after processing with the grayscale average algorithm can be expressed as:
\begin{equation}
    \bar{S}_{\text{SNR}} = \frac{\sum\limits_{x=1}^{e} \sum\limits_{y=1}^{f} {I'_0}^2(x, y)}{\sum\limits_{x=1}^{e} \sum\limits_{y=1}^{f} \left[ \frac{1}{N} \sum_{i=1}^{N} \varepsilon_i(x, y) \right]^2}=N^2\text{SNR}
    \label{eq13}
\end{equation}

Consequently, the grayscale average algorithm not only significantly improves the SNR of the processed image, which effectively enhances image quality, but also attenuates random errors introduced by heat haze.  The framework of the image restoration is introduced in Algorithm \ref{alg2}.

The processing flow of the thermal deformation image is roughly as follows: firstly, the image exposure is adjusted and the image contrast is enhanced by the multi-exposure image fusion algorithm to suppress the thermal radiation; then, the corrected image is processed by the image restoration algorithm to weaken the random noise caused by heat haze.

\section{Experiments and Results}\label{sec4}

This section evaluates the performance of the proposed algorithms through both qualitative and quantitative analyses, with a focus on their effectiveness in suppressing thermal radiation and heat haze. 

\subsection{The experimental setup}
\label{subsec41}

\begin{figure}[t]
\centering
\includegraphics[width=0.7\textwidth]{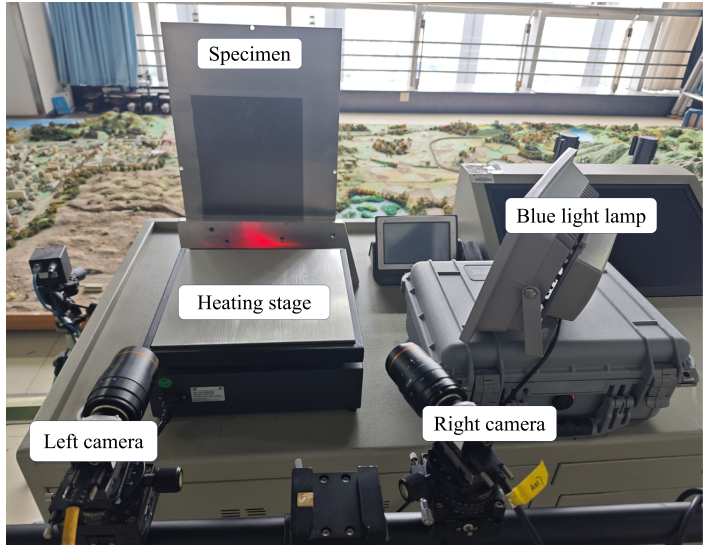}
\caption{The experimental setup of the stereo system}
\label{fig4}
\end{figure}

The experimental setup is illustrated in Fig. \ref{fig4}. A stereo imaging system is employed, consisting of two IDS cameras rigidly mounted on a tripod. The image resolution is $2048 \times 2048$ pixels, and the lenses have a focal length of 50 mm. A blue-light lamp was used, and narrow bandpass filters with a center wavelength of $450 \pm  10$ nm were mounted in front of the left and right camera lenses to suppress thermal radiation. A heating stage capable of reaching up to 450°C was placed between the stereo system and the specimen to generate heat haze. A random speckle pattern with a nominal dot diameter of 2 mm was applied to the specimen surface using water-transfer decals. Besides, active cooling via fans was employed between the heating stage and the camera setup to reduce convective heat transfer and stabilize the local environment around the optics. The calibration results of the stereo system are presented in Table \ref{tab1}. 

\begin{table}[t]
\caption{Stereo calibration results}\label{tab1}
\begin{tabular*}{\textwidth}{@{\extracolsep{\fill}}c@{\extracolsep{\fill}}c@{\extracolsep{\fill}}c@{}}
\toprule
\multicolumn{1}{c}{Parameter} & \multicolumn{1}{c}{Left camera} & \multicolumn{1}{c}{Right camera} \\
\midrule
$(f_u, f_v)$/pixel & (9767.3, 9768.6) & (9775.5, 9772.0) \\
$(u_0, v_0)$/pixel & (980.5, 990.6) & (920.8, 985.0) \\
$(k_1, k_2)$/pixel & (0.448, -0.146) & (0.418, 2.017) \\
$(p_1, p_2)$/pixel & (-0.001, -0.002) & (0.001, -0.007) \\
\multicolumn{1}{c}{$\textbf{R}$/°} & \multicolumn{2}{c}{$(1.677, 20.578, 0.477)$} \\
\multicolumn{1}{c}{$\textbf{t}$/mm} & \multicolumn{2}{c}{$(-393.8, 38.2, 55.4)$} \\
\botrule
\end{tabular*}
\footnotetext{Note: $k_1$ and $k_2$ are radial distortion coefficients; $p_1$ and $p_2$ are tangential distortion coefficients; $(f_u, f_v)$ denote the focal lengths in pixel units along the $u$- and $v$-axes; $(u_0, v_0)$ represent the principal point coordinates; and $\mathbf{R}$ and $\mathbf{t}$ are the rotation matrix and translation vector}
\end{table}

\begin{figure}[t]
\centering
\includegraphics[width=1\textwidth]{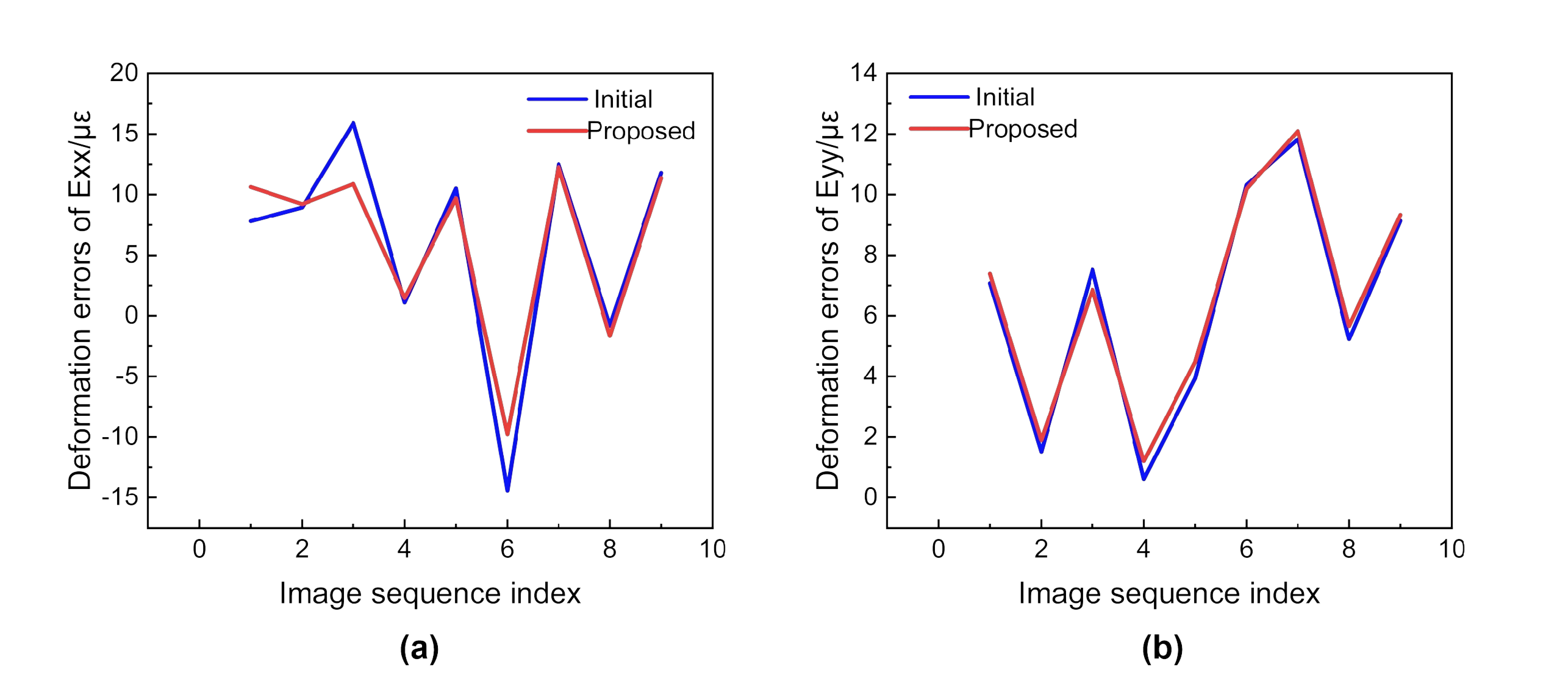}
\caption{Comparison of static noise before and after processing. \textbf{(a)} The $ \varepsilon_{xx} $ field, and \textbf{(b)} $ \varepsilon_{yy} $ field}
\label{fig8}
\end{figure}

Following calibration of the stereo system, a total of $10$ static images were acquired. The first image was used as the reference image, while the remaining $9$ served as deformation images to assess the system's static noise. The proposed multi-exposure image fusion algorithm was then applied to the entire image sequence to assess its impact on deformation measurement precision. Specifically, the subset size is set to 21 pixels, the grid step to 5 pixels, the correlation coefficient threshold to 0.8, the maximum number of iterations to 10, and both the integer search radius and stereo search radius are configured as 15 pixels. Results are shown in Fig. \ref{fig8}. In this experiment, for $\varepsilon_{xx}$ , the mean strain computed from the original images was 5.9\,\si{\micro\straina}\ (std: 9.3\,\si{\micro\straina}), and decreased slightly to 5.8\,\si{\micro\straina}\ (std: 7.6\,\si{\micro\straina}) after image enhancement. For $\varepsilon_{yy}$, the mean strain was 6.4\,\si{\micro\straina}\ (std: 3.9\,\si{\micro\straina}) before processing, and remained nearly unchanged at 6.6\,\si{\micro\straina}\ (std: 3.7\,\si{\micro\straina}) after enhancement. The mean values of both principal strain components remain below 7\,\si{\micro\straina}, indicating a low level of static noise and confirming the reliability of the experiment. Crucially, the near-constant mean strain values demonstrate that the proposed image enhancement algorithm does not compromise measurement accuracy.

To verify the effectiveness of the proposed algorithm, we simulated thermal loading experiments through system design. In the experiments, the high-power blue light lamp was used to simulate thermal radiation, and a heating stage was employed to generate heat haze. Firstly, a static image was captured under ambient temperature and used as the reference image. In addition, we collect degraded images by adjusting the power of the blue light lamp and the exposure time of the stereo system, while activating the heating stage and raising it to the highest temperature. As a result, thermal radiation and heat haze are introduced at the same time. This heat haze introduces optical aberrations, resulting in spurious strain fields computed by DIC, as shown in Fig. \ref{fig3}. While maintaining the heating stage at a constant temperature, 120 distorted images were acquired.

\subsection{Thermal Radiation Suppression Experiment}
\label{subsec42}

In the preliminary experiments, we found that although a blue-light lamp and the corresponding narrow bandpass filters were employed to suppress thermal radiation, the captured deformation images still suffered from saturation as the temperature increased. As discussed in Section \ref{subsec21}, to capture more informative images, the system's exposure time was dynamically adjusted. However, due to camera response limitations, some images exhibited overexposure or underexposure. To address this, the proposed multi-exposure image fusion algorithm was applied to correct both types of images. 

 \begin{figure}[htb]
\centering
\includegraphics[width=0.7\textwidth]{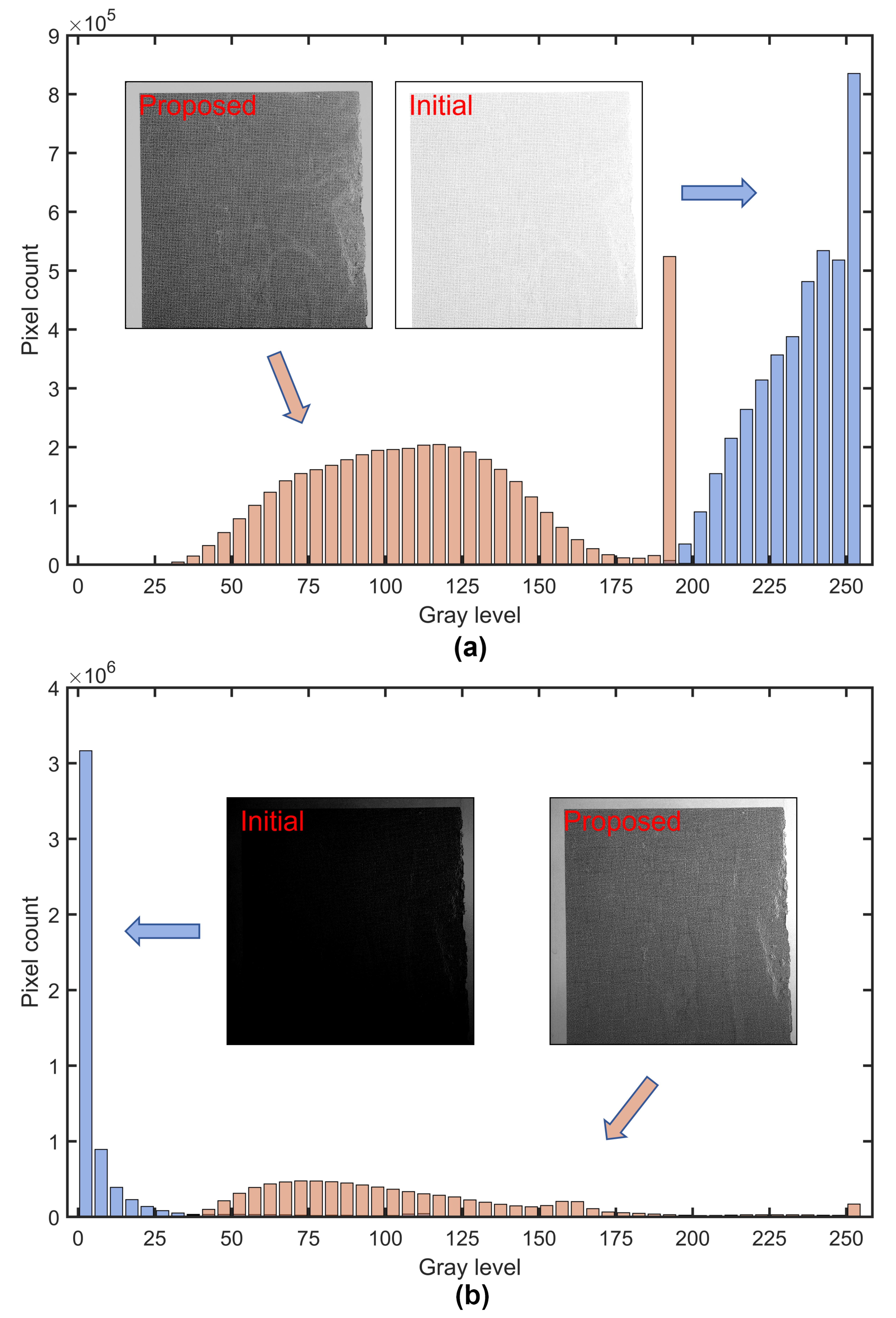}
\caption{Comparison of grayscale histograms before and after processing. \textbf{(a)} Original underexposed image and processing result of underexposed image, \textbf{(b)} original overexposed image and processing result of overexposed image}
\label{fig5}
\end{figure}

To evaluate the effectiveness of the proposed algorithm, an underexposed and an overexposed image captured by the left camera were selected for processing. The speckle pattern quality before and after enhancement was assessed using grayscale histograms. Generally, high-quality speckle images are characterized by broad and relatively uniform grayscale distributions. Figure \ref{fig5} demonstrates that, after processing with our proposed algorithm, both underexposed and overexposed images achieve improved contrast and enhanced speckle texture clarity.

In addition, we quantitatively evaluated the underexposed and overexposed images before and after the algorithm processing through image quality indicators and effective deformation calculation area. Approximately 60 image pairs were analyzed, with the results averaged to assess the overall performance of the proposed algorithm. The image quality indicators employed include the no-reference image quality metric for contrast distortion (NIQMC) \cite{Gu2017}, natural image quality evaluator (NIQE) \cite{Gu2015}, and mean intensity gradient (MIG) \cite{PAN2010}. For NIQMC, higher scores correspond to superior contrast quality. And a lower NIQE score corresponds to superior perceptual image quality, as the metric quantifies deviations from statistical regularities. MIG is particularly used for the quality assessment of speckle patterns, with a high-quality speckle pattern distinguished by a large mean intensity gradient. Table~\ref{tab2} presents the quantitative image quality assessment results, with red entries indicating superior performance under the respective evaluation criteria. As shown, for both underexposed and overexposed images, image quality is significantly improved after processing.

 \begin{figure}[t]
\centering
\includegraphics[width=1\textwidth]{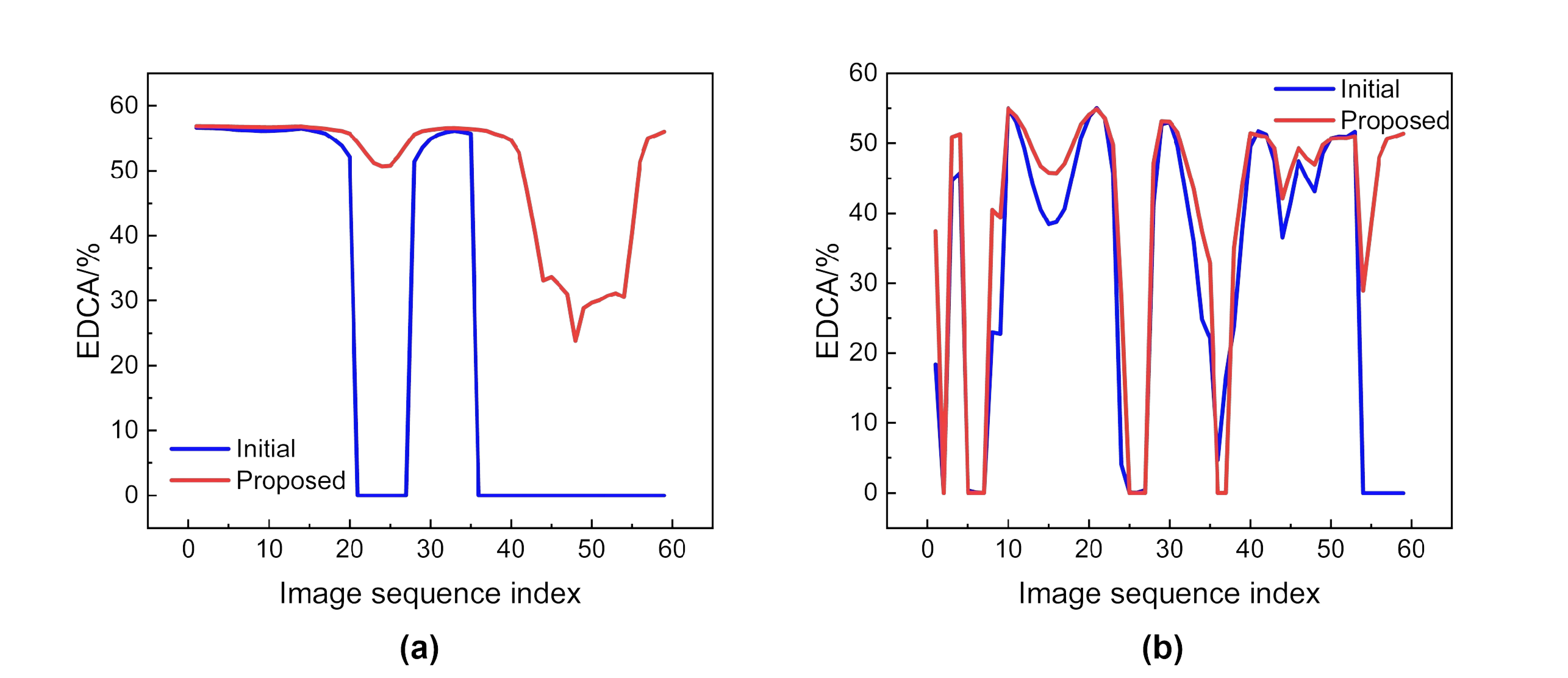}
\caption{Comparison of EDCA before and after processing. \textbf{(a)} Statistical results of EDCA for underexposed images, and \textbf{(b)} statistical results of EDCA for overexposed images}
\label{fig6}
\end{figure}

\begin{table}[t]
  \centering
  \caption{Comparison of image quality indicators before and after processing}
  \begin{tabular*}{\linewidth}{@{\extracolsep{\fill}}lcccc@{}}
    \toprule
    Exposure & Method & NIQMC$^{\uparrow}$ \cite{Gu2017} & NIQE$^{\downarrow}$ \cite{Gu2015} & MIG$^{\uparrow}$ \cite{PAN2010} \\
    \midrule
    \multirow{2}{*}{Underexposed}
             & Initial  & $0.09$ & $4.87$ & $0.75$ \\
             & Proposed & $\textcolor{red}{4.54}$ & $\textcolor{red}{3.16}$ & $\textcolor{red}{5.28}$ \\
    \midrule
    \multirow{2}{*}{Overexposed}
             & Initial  & $2.79$ & $5.96$ & $5.61$ \\
             & Proposed & $\textcolor{red}{3.58}$ & $\textcolor{red}{3.54}$ & $\textcolor{red}{9.34}$ \\
    \bottomrule
  \end{tabular*}
  \label{tab2}
  \footnotetext{Note: $\uparrow$ indicates that larger values values are better,while $\downarrow$ indicates that smaller values are better}
\end{table}

The effective deformation calculation area (EDCA) is defined as the percentage of image pixels for which a reliable full-field strain value can be successfully computed by the DIC algorithm, expressed as a percentage (\%). To facilitate quantitative analysis, statistical evaluation was performed on the aforementioned 60 image pairs. As shown in Fig. \ref{fig6}, under underexposure conditions, the EDCA of the original images is 26.40\% (std: 28.03\%), and after multi-exposure image fusion, it increases to 50.19\% (std: 10.16\%). Under overexposure conditions, the mean EDCA rises from 32.14\% (std: 20.72\%) to 40.00\% (std: 18.12\%). These results indicate that the proposed multi-exposure image fusion method significantly improves both data completeness and measurement stability. Furthermore, Figure \ref{fig6} indicates that the proposed algorithm is more effective for underexposed images. This is because underexposed images typically retain the underlying speckle texture and sample details, albeit with low contrast that hinders visibility. In contrast, overexposed images suffer from severe pixel saturation, leading to irreversible loss of structural information in critical regions. 

\subsection{Heat Haze Suppression Experiment}
\label{subsec43}
Once the temperature of the heating stage reached 450°C, it was held constant, and images were continuously acquired using the stereo imaging system to perform static thermal loading experiments. Note that no mechanical or thermal loading was applied to the specimen throughout the experiment. Therefore, the true strain remained zero. The deformation fields computed by DIC are thus attributed entirely to measurement errors induced by heat haze, providing a quantitative evaluation of the algorithm's effectiveness. 

After processing the distorted images with the multi-exposure fusion algorithm, to enable a comprehensive comparative analysis, three different processing strategies were applied to the acquired images: (1) the fused images, (2) the fused images processed using the grayscale average algorithm, and (3) the fused images processed by our image restoration algorithm.  The corresponding deformation fields were computed for each case to evaluate the effectiveness of the proposed method. Note that the grayscale average algorithm was implemented using a sequence of 15 deformation images.

\begin{table}[t]
  \centering
  \caption{Strain measurements by different strategies}
  \begin{tabular*}{\linewidth}{@{\extracolsep{\fill}}lccc@{}}
    \toprule
    Method & $ \varepsilon_{xx} $/ \si{\micro\straina} & $ \varepsilon_{yy} $/ \si{\micro\straina} & $ \gamma_{xy} $/ \si{\micro\straina} \\
    \midrule
    Fused    & $232$             & $211 $              & $22$            \\
    Average \cite{SU2015}   & $34 $              & $135$               & $-16$            \\
    Proposed    & $\textcolor{red}{28}$   & $\textcolor{red}{134}$         & $\textcolor{red}{-14}$            \\
    \bottomrule
  \end{tabular*}
  \label{tab4}
  \footnotetext{Note: \textit{Fused} refers to the raw multi-exposure fused image, \textit{Average} denotes the fused image further processed by grayscale averaging, and \textit{Proposed} corresponds to the fused image enhanced solely by our FSIM-guided restoration algorithm—without incorporating the grayscale averaging step}
\end{table}

The experimental results are summarized in Table \ref{tab4}, where the minimum error values across the three strategies for each strain component are highlighted in red. As shown, the image restoration algorithm significantly reduces measurement errors induced by heat haze. Specifically, for $\varepsilon_{xx}$, the error decreases from 232\,\si{\micro\straina}\ to 34\,\si{\micro\straina}, representing a reduction of 85.3\%; for $\varepsilon_{yy}$, the error is reduced from 211\,\si{\micro\straina}\ to 135\,\si{\micro\straina}, a 36.0\% improvement; and for $\gamma_{xy}$, the absolute error decreases from 22\,\si{\micro\straina}\ to 14\,\si{\micro\straina}, corresponding to a 36.4\% reduction. These results demonstrate the effectiveness of the image restoration in mitigating heat haze in high-temperature DIC measurements.

The above experiments demonstrate the effectiveness of the proposed image processing algorithms. The multi-exposure image fusion algorithm significantly improves image contrast and overall quality, thereby increasing the valid measurement area over which reliable deformation fields can be computed using DIC. Importantly, the results confirm that the image enhancement process does not compromise measurement accuracy, as no significant bias is introduced into the deformation calculations. Meanwhile, the image restoration algorithm effectively reduces stochastic errors caused by heat haze, suppressing strain fluctuations. However, this method is only applicable to static or quasi-static thermal deformation tests, as it relies on temporal averaging over multiple images and is therefore unsuitable for dynamic deformation scenarios.

\section{Conclusion}\label{sec5}

\section*{Conclusion}
This study presents an image processing framework to mitigate the adverse effects of thermal radiation and heat haze in high-temperature deformation measurements using DIC. These key challenges are addressed through a combination of multi-exposure image fusion and image restoration algorithms.

A multi-exposure image fusion algorithm is proposed to suppress overexposure and underexposure caused by complex illumination conditions at elevated temperatures. Experimental results demonstrate that this method effectively improves image contrast and texture clarity, thereby expanding the EDCA. Notably, the algorithm introduces no measurable bias in the computed strain fields. Furthermore, to mitigate random noise caused by heat haze, an image restoration algorithm is deployed. Specifically, the disturbance parameters are corrected by maximizing the FSIM between the reference image and the target image as the objective function, thereby realizing image correction. Subsequently, a grayscale averaging algorithm is implemented on a sequence of 15 deformation images to further suppress random errors. 

In summary, the proposed image processing strategy effectively improves image quality and reduces deformation measurement errors under high-temperature conditions. Given that the framework relies on grayscale averaging, the overall proposed framework is applicable only to static or quasi-static thermal loading experiments. Despite this limitation, the framework still exhibits strong potential for application in high-temperature structural testing, thermal expansion characterization, and other fields that demand reliable full-field deformation data in thermally harsh environments.

\bmhead{Acknowledgements}

This work is supported by the Hunan Provincial Natural Science Foundation for Excellent Young Scholars (Grant 2023JJ20045) and the National Natural Science Foundation of China (Grant 12372189). 

\section*{Declarations}
\textbf{Conflict of interest}   The authors declare that they have no conflict of interest.

\bibliography{sn-bibliography.bib}

\end{document}